\documentclass[letterpaper, 10 pt, conference]{ieeeconf}  

\IEEEoverridecommandlockouts                              

\usepackage{graphics} 
\usepackage{epsfig} 

\usepackage{hyperref}
\hypersetup{
    colorlinks=true,
    linkcolor=blue,
    urlcolor=blue,
    citecolor=blue,
    pdftitle={SLAMFuse},
    pdfpagemode=FullScreen,
    }
\usepackage{tabularx}
\newcolumntype{C}{>{\centering\arraybackslash}X}
\usepackage{gensymb}
\usepackage{multirow, array}
\usepackage{caption}
\usepackage{subcaption}
\title{\LARGE \bf A Framework for Reproducible Benchmarking and Performance Diagnosis of SLAM Systems}

\author{Nikola Radulov$^{1*}$, Yuhao Zhang$^{1*}$, Mihai Bujanca$^2$, Ruiqi Ye$^1$, Mikel Luj\'an$^1$\\
\thanks{$^{*}$ Equal Contribution}%
\thanks{$^{1}$ Department of Computer Science, University of Manchester, UK.}%
\thanks{$^{2}$ Qualcomm Technologies XR Labs, Austria.}}%

\begin{document}

\maketitle
\pagestyle{empty}

\begin{abstract}
We propose SLAMFuse, an open-source SLAM benchmarking framework that provides consistent cross-platform environments for evaluating multi-modal SLAM algorithms, along with tools for data fuzzing, failure detection, and diagnosis across different datasets.
Our framework introduces a fuzzing mechanism to test the resilience of SLAM algorithms against dataset perturbations. This enables the assessment of pose estimation accuracy under varying conditions and identifies critical perturbation thresholds. SLAMFuse improves diagnostics with failure detection and analysis tools, examining algorithm behaviour against dataset characteristics.
SLAMFuse uses Docker to ensure reproducible testing conditions across diverse datasets and systems by streamlining dependency management. Emphasizing the importance of reproducibility and introducing advanced tools for algorithm evaluation and performance diagnosis, our work sets a new precedent for reliable benchmarking of SLAM systems.
We provide ready-to-use docker compatible versions of the algorithms and datasets used in the experiments, together with guidelines for integrating and benchmarking new algorithms.
Code is available at \href{https://github.com/nikolaradulov/slamfuse}{https://github.com/nikolaradulov/slamfuse}

\end{abstract}

\begin{figure*}[!ht]
    \centering
    \includegraphics[width=0.99\textwidth]{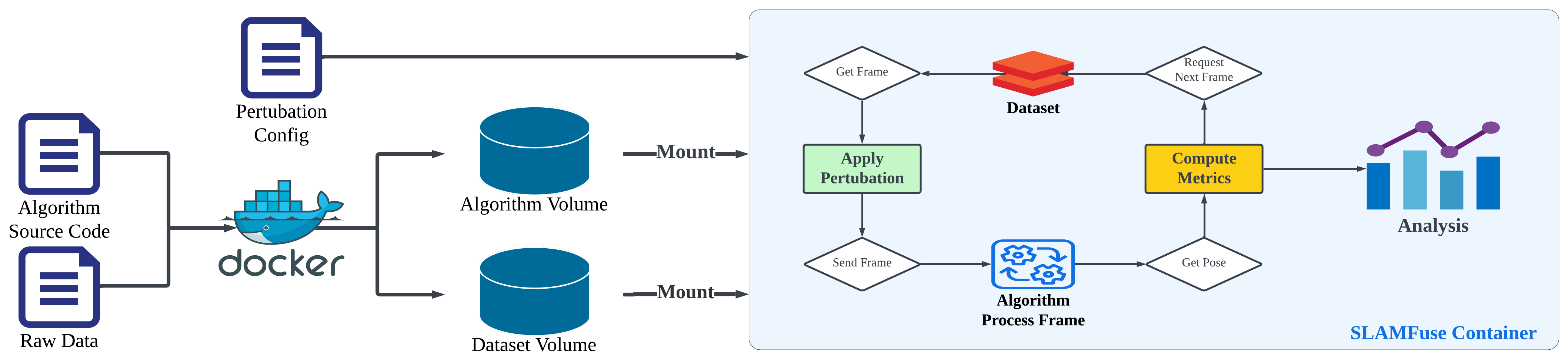}
    \caption{Architecture the SLAMFuse benchmarking framework: The source code of the algorithm and raw dataset are encapsulated in separate Docker volumes, which are mounted into the Docker container of SLAMFuse. SLAMFuse then takes frames from the dataset volume, applies perturbations, and passes them to the algorithm. Different metrics are computed simultaneously for analysis.}
    \label{fig:slambench-pipeline}
\end{figure*}



\section{INTRODUCTION}
SLAM (Simultaneous Localization and Mapping) is a set of algorithms used by robots and autonomous vehicles to create maps of their surroundings while simultaneously determining their pose within the maps. SLAM finds applications in areas such as hazardous environment exploration, search and rescue, and autonomous vehicles operating in challenging environments. This has led to the adoption of a wide range of sensors \cite{sensors} in SLAM implementations either as alternatives or as complements to the existing systems. For example, Light Detection And Ranging (LiDAR) sensors offer unparalleled accuracy and the ability to capture details from the surrounding environment, even in challenging lighting conditions and areas with obstacles. This makes LiDAR sensors particularly valuable for applications demanding high-resolution spatial data. Unsurprisingly, each sensor provides unique advantages and disadvantages, with choices often dictated by the specific deployment requirements of an application, such as cost, precision, and efficiency.

To help researchers analyze and compare SLAM systems, a framework is needed to accommodate a wide range of sensors and evaluate them across datasets. To that end, we introduce SLAMFuse, a framework that supports a diverse array of sensors and datasets, allowing the evaluation of SLAM systems using qualitative and quantitative metrics, with plug-and-play algorithm support. By extending \cite{slam3}, our framework can directly compare the performance of algorithms with different sensors under a plethora of environmental conditions through the wide range of supported datasets. Having such a comprehensive benchmark suite not only removes the guesswork from selecting the right algorithm for a particular scenario, but also helps in identifying the limitations of SLAM algorithms by exposing correlations between input data and failures, or increases in error.


Our contributions can be summarised as follows:
\begin{itemize}
    \item A multi-modal benchmarking framework for SLAM algorithms using LiDAR, IMU, and traditional cameras, including monocular, stereo, and RGB-D cameras.
    \item A fuzzing mechanism for understanding the operating bounds of SLAM algorithms.
    \item A failure detection and analysis tool that identifies correlations between visual input quality and SLAM algorithm errors that can aid in identifying algorithm weaknesses.
    \item The first SLAM benchmarking framework utilises both Docker containers and volumes to create self-contained, reproducible environments across datasets and computing platforms.
\end{itemize}

\section{Background and Related Work}
\subsection{Visual SLAM}
The goal of visual SLAM is to extract information from data captured by the visual sensors to create a map of the environment while estimating the sensor's translation and rotation within that map. Visual SLAM systems may also use IMU and depth information to achieve better results, however, traditional approaches make use of frame-based cameras, such as rolling or global shutter cameras, which are sensitive to motion blur, and poor illumination and are subject to low dynamic range. On the other hand, LiDAR is not sensitive to illumination conditions \cite{lidar}. 

\subsection{LiDAR-based SLAM}
\label{section::lidar}
LiDAR-based SLAM utilizes sensors which rely on laser pulses to measure distances to nearby objects with high accuracy. Similar to RGB-D-based SLAM systems, it is also possible for LiDAR SLAMs to construct a dense 3D map of the surrounding environment. The ability of LiDAR sensors to provide detailed and precise spatial information despite lighting conditions makes them an essential sensor in the development and implementation of state-of-the-art SLAM systems.

LOAM (LiDAR Odometry and Mapping) \cite{loam} is a widely recognized LiDAR-based SLAM system known for its ability to perform real-time localization and mapping by matching feature edges and planes. Building on LOAM, F-LOAM \cite{floam} offers an faster open-source version. LeGO-LOAM \cite{legoloam} enhances capabilities for ground vehicles, delivering efficient pose estimation in challenging terrains through advanced point cloud segmentation and optimization techniques.

\subsection{SLAM Benchmarking Tools}
Several benchmarks such as the KITTI \cite{kitti_benchmark} and TUM RGB-D \cite{tum_benchmark} benchmarking suites have been proposed with the aim to assess the accuracy, efficiency, and robustness of different SLAM algorithms. 

The SLAMBench framework \cite{slam1} has iteratively gone above and beyond simple benchmarking tools. SLAMBench2 \cite{slam2} was the first framework to integrate a variety of algorithms and datasets into the necessary tools to compare SLAM systems. SLAMBench3 \cite{slam3} further developed support for dynamic and semantic SLAM, but did not extend the range of available sensors beyond cameras and IMU. Further efforts \cite{bujanca2021robust} examined the robustness of SLAM algorithms with respect to perturbations inherent in datasets, such as illumination changes and dynamic objects. Building upon the previous SLAMBench efforts, we address this limitation in our framework by introducing LiDAR sensor support. 

Building SLAM algorithms is not trivial due to dependency conflicts and compatibility issues, thus making SLAM benchmarking difficult. Docker \cite{docker}, as a tool for creating and managing containerized applications, is widely used in SLAM algorithms for its capability to address compatibility issues. For example, algorithms such as,  \cite{orbslam2} \cite{orbslam3} \cite{legoloam}, already have their docker version. Although running SLAM algorithms in a single isolated Docker container is straightforward, adapting this approach to a large framework, such as a benchmarking platform where multiple isolated building environments are needed and data sharing is required,  continues to be a challenge.
While SLAM Hive \cite{slam-hive} is the first SLAM system evaluation framework that makes use of Docker containers, we move beyond this by harnessing Docker volumes. In addition, we also conduct the first experiments to quantify the run-time overhead of executing SLAM algorithms under Docker.
By deconstructing SLAMBench into separate components and reassembling them using Docker, our framework is the first one to address library dependency issues among algorithms. Thus, allowing for greater portability and making the integration of future algorithms easier.

\section{Methodology}
SLAMFuse provides new tools that not only facilitate the interpretation of SLAM algorithm behaviour but also expand multi-modal support. We construct a fuzzing-based system that eases the understanding of the relationship between input data and the algorithm performance.
Identifying shortcomings in \cite{slam1} \cite{slam2} \cite{slam3}, we begin to reconsider the architecture with a focus on enhancing framework maintainability, algorithm portability, and user accessibility. Software engineering challenges, such as library dependency conflicts among algorithms, push the framework to embrace containerization. The aim is seamless ``plug-and-play" integration and usage of SLAM systems and datasets. Figure \ref{fig:slambench-pipeline} shows the Dockerized architecture of the SLAMFuse framework. Our framework and integrated algorithms can be easily built without any modifications to the source code on several platforms, including different versions of Ubuntu, and Windows Subsystem for Linux.

\subsection{Container-based Architecture}
In the absence of an isolated, container-based environment, building dependencies for two or more SLAM systems is challenging due to the lack of compatibility across their libraries. For example, we could not find a compatible version of the Eigen library for both ORB-SLAM2 \cite{orbslam2} and ORB-SLAM3 \cite{orbslam3} on Arm processors. To address this problem, we have developed a new solution that leverages Docker containers and volumes. This approach allows each algorithm to work as a plug-and-play entity.

\subsubsection{SLAMFuse Container} 
The SLAMFuse Container represents a lightweight, standalone, and executable software package that includes the tools needed to run a benchmarking experiment using an algorithm, a dataset, and appropriate libraries. Essentially, the data loaders and the main compute loop existing in SLAMFuse are now packed inside a docker container. To run an experiment, an \texttt{Algorithm} and a \texttt{Dataset} are made available to the SLAMFuse container through mounted volumes. Each algorithm is packed with its libraries into one API volume. Thus, the SLAMFuse container requires at least one dataset and one SLAM algorithm volume to run.  More complex experiments would involve one SLAMFuse container with multiple dataset volumes mounted and multiple API volumes. Each API volume is designed to prevent dependency conflicts with other algorithms by allowing the algorithm to locate its necessary library dependencies within its designated API volumes while it executes. This approach provides an isolated and consistent experimental environment, ensuring that the algorithm behaves the same regardless of the underlying architecture.


\subsubsection{Algorithm Volume}
Each algorithm requires a Dockerfile that specifies the steps to compile the dependencies and the algorithm itself. The image that is built from the Dockerfile is then used to populate the volume containing the essential header files and libraries for the dependencies. 

\subsubsection{Dataset Volume}
To ensure experiments remain efficient, we opted to separate the datasets from the SLAMFuse container. This approach enables a given experiment to mount and download only the required datasets. We continue encapsulating the datasets in \texttt{.slam} files as in \cite{slam2}, which organize all necessary sensor data for benchmark processing. This \texttt{.slam} file consists of serialized instances of the \texttt{Frame} class, each equipped with attributes specifying the time stamp, sensor type, and corresponding data, among other details.



\begin{table}[!ht]
    \centering
    \caption{Datasets used in Experiments.}
    \begin{tabular}{|c|c|}
        \hline
        Datasets & Sensors \\
        \hline
        TUM RGB-D \cite{tum_benchmark} & RGB-D \\
        KITTI \cite{kitti_raw_data} & LiDAR, Stereo, IMU \\
        Newer College \cite{newercollege} &  LiDAR, Stereo, IMU \\
        DARPA Subterranean Challenge \cite{darpa2} &  LiDAR, Stereo, IMU \\
        UZH-FPV \cite{uzhfpv} & Stereo, IMU, Event Camera \\
        \hline
    \end{tabular}
    \label{tab:datasets}
\end{table}
\begin{table}[!ht]
    \centering
    \caption{SLAM Algorithms used in the experiments.}
    \begin{tabular}{|c|c|c|}
        \hline
        Algorithms & Type & Sensors \\
        \hline
        ORB-SLAM3 \cite{orbslam3}  & Sparse & RGB-D, Stereo, Mono, IMU \\
        LSD-SLAM \cite{lsdslam} & Semi-Dense & Mono \\        ElasticFusion \cite{elasticfusion} & Dense & RGB-D \\
        KISS-ICP \cite{kissicp} & Semi-Dense & LiDAR \\
        LOAM \cite{loam} & Semi-Dense & LiDAR \\
        F-LOAM \cite{floam} & Semi-Dense & LiDAR \\
        LeGO-LOAM \cite{legoloam} & Semi-Dense & LiDAR, IMU \\
        \hline
        \noalign{\vskip 1mm}
    \end{tabular}
    \label{tab:algorithms}
\end{table}

\subsection{Fuzzing SLAM -- Adding Perturbations with Control} \label{section:fuzzing}
Fuzzing, a widely used technique in software engineering, involves feeding unexpected data to a program to analyse its resilience. Inspired by this practice, we propose a fuzzing mechanism tailored for SLAM algorithms. This mechanism enables the introduction of perturbations into datasets, which are then supplied to the algorithms. Notably, these changes occur dynamically at runtime, coinciding with the introduction of each frame to the algorithm (see Figure~\ref{fig:slambench-pipeline}) without necessitating additional storage for the modified datasets. Furthermore, there is no discernible overhead on the algorithm under stress, as the modifications are performed external to the processing and hence do not contribute to the measured execution time of the algorithm.

To ensure reproducibility, we implement configuration files in JSON format, specifying the characteristics of the perturbations to be applied. These configurations encompass various aspects, including specifying frames from which sensors should be modified, identifying frames to be modified by ID, and defining settings for different perturbations to be applied. Importantly, these configurations can be customized on a per-frame basis, ensuring flexibility and adaptability to diverse scenarios.


\subsection{Image Metrics}
Beyond evaluating trajectory error, map accuracy, and computational efficiency, we also incorporate metrics to assess image quality for visual SLAM. This includes brightness, contrast, and sharpness. Brightness and contrast are determined by the mean and standard deviation of the pixel intensity levels. For assessing sharpness and blur, we employ the Tenengrad method with the Sobel operator and normalise the outcome by the dimensions of the image, as illustrated in Equation~\ref{eq:tenengrad}. A lower value indicates greater blur.

\begin{equation}
T = \frac{\sum_{x, y} \sqrt{(Sobel_x(x, y)^2 + Sobel_y(x, y)^2)}}{width \times height}
\label{eq:tenengrad}
\end{equation}
\vspace{1pt}

\subsection{Performance Diagnostics Tool}\label{peformance}
One of the core innovations of our framework is the performance diagnostics tool, which can be used to gain insights into the performance of an algorithm on new data.
Since our framework offers full control over the execution flow of the integrated SLAM algorithm, we can compute metrics for each frame and autonomously detect frames with failure or near-failure situations (ie. frames with significant relative pose error), as well as restart the algorithm from these pinpointed frames. Importantly, we also allow the user to define their own failure parameters, which are based on error levels, reported pose, and execution time.
This tool can be used to understand the algorithm performance on difficult datasets, where failures or significant degradation might occur. 
As before, the error and input metrics are collected when running the algorithm. The diagnostics tool processes this data and selects frames of interest by isolating frames where the error exceeds a threshold set by the user, then selects a window of frames around the failure point such as in Figures \ref{Figure:lsd-drone} and~\ref{fig:darpa images quality}. 




\begin{table*}[tb!]
\caption{Docker Overhead on Windows Linux Subsystem and Native Ubuntu}\label{tab:dockeroverhead}
\centering
\resizebox{\linewidth}{!}{%
\begin{tabular}{|c|c|c|c|>{\bfseries}c|c|c|>{\bfseries}c|}
\hline
Datasets & Algorithms &  WSL $^\mathbf{d}$ & WSL & Overhead WSL & Ubuntu $^\mathbf{d}$& Ubuntu & Overhead Ubuntu\\
\hline
\multirow{2}{*}{KITTI} 
& LOAM & $3.94 \pm 0.06$ &$4.00 \pm 0.20$& 1.31\% & $4.24 \pm  0.04 $& $4.47 \pm 0.06$& 5.14\% \\
& LeGO-LOAM &  $7.86\pm0.07$ & $7.97 \pm 0.07 $ & 1.46\% &  $7.76 \pm 0.04$ &$8.14 \pm 0.02$ & 4.67\%\\
\hline
\multirow{2}{*}{Newer College}
& LOAM & $2.94\pm 0.02$ & $3.04 \pm 0.10$ & 3.51\% &$3.33\pm 0.07$ & $3.60 \pm 0.05$ & 7.5\%\\
& LeGO-LOAM & $5.47 \pm 0.11$ & $5.52 \pm 0.17$ & 0.91\% & $5.15\pm 0.13$ & $5.74 \pm 0.01$ & 10.21\%\\
\hline
\noalign{\vskip 1mm}
\multicolumn{7}{l}{$^\mathbf{*}$Performance is measure in FPS. Algorithms are modified to run sequentially to eliminate the non-determinism introduced by multi-threading.}\\
\multicolumn{4}{l}{$^\mathbf{d}$algorithms running with docker.}\\
\end{tabular}
}

\label{Table:overhead—docker}
\end{table*}

\subsection{Multi-Modality}
Different sensors have distinct advantages and disadvantages, making SLAM algorithms based on various modalities suited for different use cases. To facilitate this, SLAMFuse supports different types of sensors and integrates open-source state-of-the-art algorithms from different modalities (see Table~\ref{tab:algorithms}), along with datasets that include various sensors (see Table~\ref{tab:datasets}). In addition, the evaluation includes a comparison of a robust visual SLAM algorithm and the added LiDAR algorithms (see Table \ref{tab:algorithms}). 

Leveraging \cite{slam2}, SLAMFuse has a class hierarchy with a simple API abstracting away the difference across sensors. LiDAR sensor support has been added by creating new subclasses. In addition, integrated datasets with different modalities are synchronized, rectified, and stored in a unified data format. For example, LiDAR data is stored in a PCL point cloud \cite{pcl}, which facilitates using the mature tools offered by the PCL software ecosystem.





\begin{figure*}[bth!]
    \centering
    \includegraphics[width=0.9\textwidth]{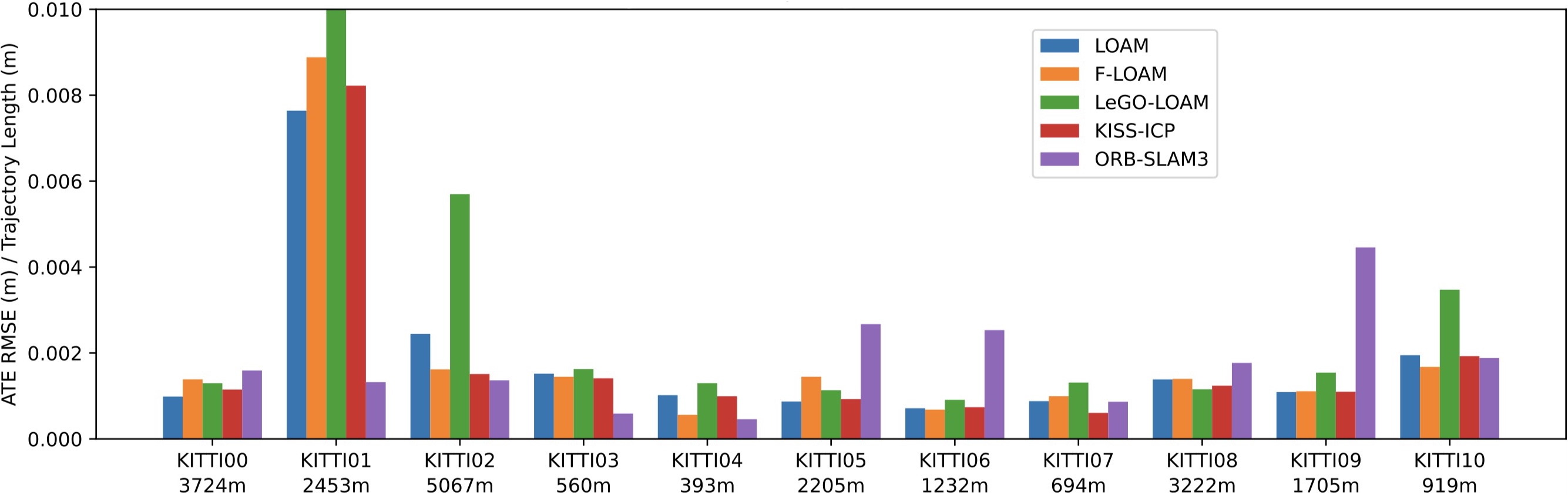}
    \caption{Comparison of ATE RMSE on KITTI sequences.}
    \label{Figure:kitti_ate_rmse}
\end{figure*}
\section{Experiments}


\subsection{Experimental Setup}
The experiments revolve around four main aspects. First, we consider the overhead introduced by using containers. Second, we evaluate algorithm performance under standard conditions using established datasets and we explore the potential benefits of LiDAR approaches in scenarios where visual SLAM algorithms may falter. The LiDAR-based algorithms include KISS-ICP \cite{kissicp}, known for its versatility and ease of use, LOAM \cite{loam}, a seminal paper in LiDAR SLAM, F-LOAM \cite{floam} and LeGO-LOAM \cite{legoloam}, extensions of LOAM emphasizing lightweight performance. For this purpose, we utilize the KITTI \cite{kitti_benchmark}, Newer College \cite{newercollege}, and DARPA Subterranean Challenge \cite{darpa2} datasets. These datasets encompass both stereo cameras and LiDAR sensors, facilitating direct comparisons between the LiDAR algorithms and ORB-SLAM3 \cite{orbslam3}. Note that for Newer College and DARPA, we provided a simplified version with all data synchronized to 10Hz.

Third, systematically perturb some of these datasets to observe how changes in error correlate with algorithmic performance degradation, effectively identifying points of failure. For our fuzzing experiments, we focus on the TUM dataset, with LSD-SLAM, ORB-SLAM3 and ElasticFusion selected as representatives of semi-dense, sparse and dense algorithms. The datasets aim to showcase the capabilities of our tool and they are deliberately challenging scenarios for these algorithms. 

Lastly, we analyse failure situations and correlate them against the mentioned metrics. We select the UZH-FPV dataset, known for its challenges in tracking fast-paced dynamic environments with illumination changes and occlusions.

The experiments are conducted on a system featuring an AMD Ryzen 9 5900HS processor, 16GB of RAM, and an NVIDIA GeForce GTX 1650 GPU, operating on both Windows 10 with WSL (Windows Subsystem for Linux) and Ubuntu 20.04 platforms.

\subsection{Overhead of using Containers}
We measure the overhead of using Docker containers compared to running directly on the host machine. We run LOAM \cite{loam} and LeGO-LOAM \cite{legoloam} with different datasets of varying difficulty. Table \ref{tab:dockeroverhead} reports the additional load as a percentage decrease of FPS when executed within the container in comparison to the measurements taken on the host system. 

On WSL, the overhead is minimal, with a maximum of 3\% being observed. In contrast, on native Ubuntu, the reported overhead is larger, typically ranging between 5\% and 10\%.

The lower overhead of Docker on Windows compared to native Ubuntu may be due to both Docker and non-Docker applications on Windows utilizing WSL2. As a result, the additional layer of complexity seen with Docker on native Ubuntu affects both Dockerized and non-Dockerized versions equally on Windows, making the overhead appear lower. Furthermore, the algorithms being 3\% to 15\% slower on Windows compared to Ubuntu might indicate that the WSL2 layer contributes to this performance degradation. Thus, the comparison might not be entirely fair, as the overhead attributed to Docker on Windows might be less pronounced due to the shared WSL2 layer. However, this explanation remains speculative based on the current data. Further experiments, such as running the benchmark natively on Windows without WSL, are needed to confirm whether WSL2 introduces significant overhead and to clarify the impact of Docker versus the WSL2 environment.



\subsection{Different Modality Evaluation}
Note this part of the evaluation does not include ElasticFusion or LSD-SLAM. This is because KITTI, Newer College and DARPA do not provide RGB-D input required for ElasticFusion, and LSD-SLAM struggles when trajectories are mainly forward movement, as acknowledged in the original LSD-SLAM GitHub repository.

\subsubsection{KITTI}
All KITTI sequences ranging from 00 to 10 are used. To standardize the results for clarity on a unified graphical representation, we normalize the ATE RMSE values (in meters) by dividing them by their respective trajectory lengths (in meters). Figure \ref{Figure:kitti_ate_rmse} presents a comparative analysis of SLAM algorithms. All LiDAR algorithms demonstrate lower-than-usual accuracy with sequence KITTI 01, which is characterized by high-speed and long-duration highway driving.
This reduced accuracy is due to a notable decrease in the number of extracted edge features, primarily from a nearby high-speed vehicle, impacting odometry and map optimization accuracy. Moreover, we observe that ORB-SLAM3 performs better on sequence 01, because the camera sensor data is not obstructed by highway guardrail, preserving the integrity of the visual information and enabling more effective operation.

\begin{table}[tb!]
\caption{Comparison of LiDAR and Visual SLAM}
\centering
\begin{tabular}{|c|c|c|} 
\hline
Datasets & Algorithms & ATE RMSE (m) \\
\hline
\multirow{5}{*}{Newer College Quad and Mid}
& LOAM      & 0.249 \\
& F-LOAM    & 0.403 \\
& LeGO-LOAM & 0.191 \\
& KISS-ICP  & 0.166 \\
& ORB-SLAM3 & 1.243 \\
\hline
\multirow{5}{*}{\parbox{3.8cm}{\centering DARPA Subterranean Challenge \\ ANYmal-1}}
& LOAM      & 1.596 \\
& F-LOAM    & 1.049 \\
& LeGO-LOAM & Fail \\
& KISS-ICP  & Fail \\
& ORB-SLAM3 & Fail \\
\hline
\end{tabular}
\label{Table:lidar_comparison}
\end{table}
\subsubsection{Newer College}
\begin{figure}[tb!]
    \centering
    \begin{subfigure}[a]{\columnwidth}
        \centering
        \includegraphics[width=0.97\linewidth]{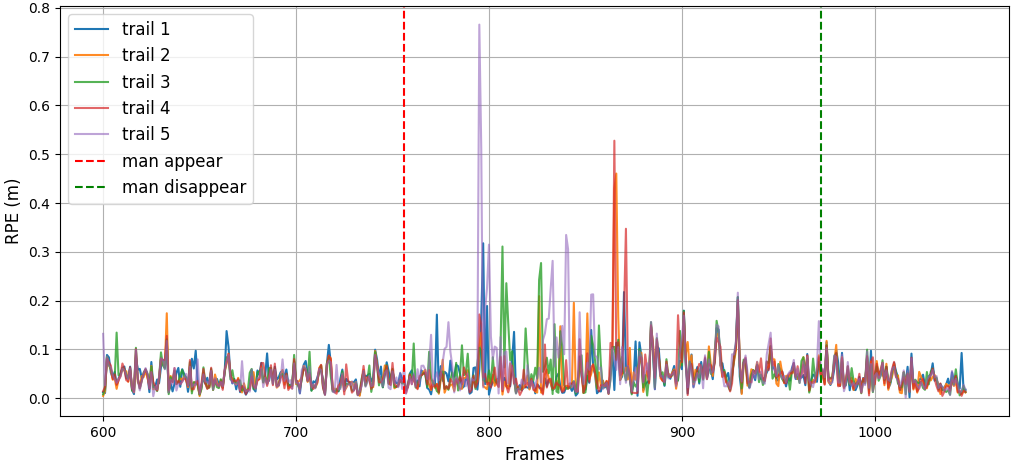}
        \caption{}
        \label{Figure:orbslam3_rpe_sub}
    \end{subfigure}

    \vspace{1ex} 
    
    \begin{subfigure}[b]{\columnwidth}
        \begin{minipage}[b]{0.49\linewidth}
            \includegraphics[width=\linewidth, height=0.55\linewidth]{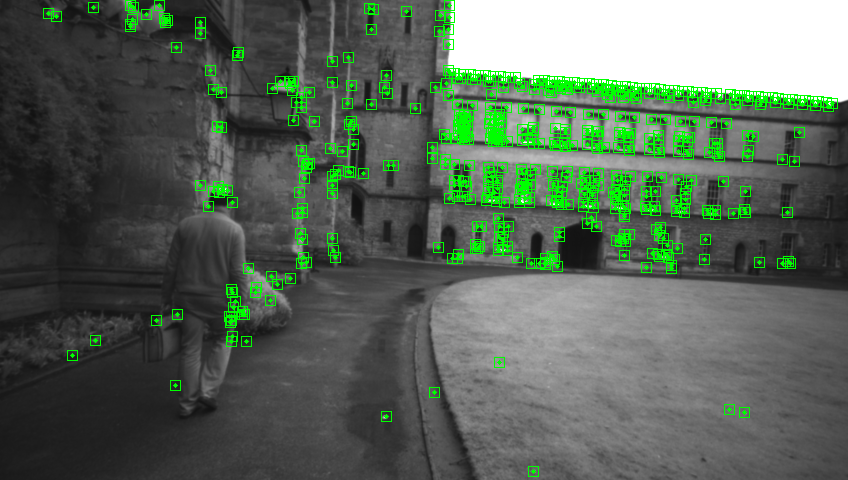}
        \end{minipage}
        \hfill
        \begin{minipage}[b]{0.49\linewidth}
            \includegraphics[width=\linewidth, height=0.55\linewidth]{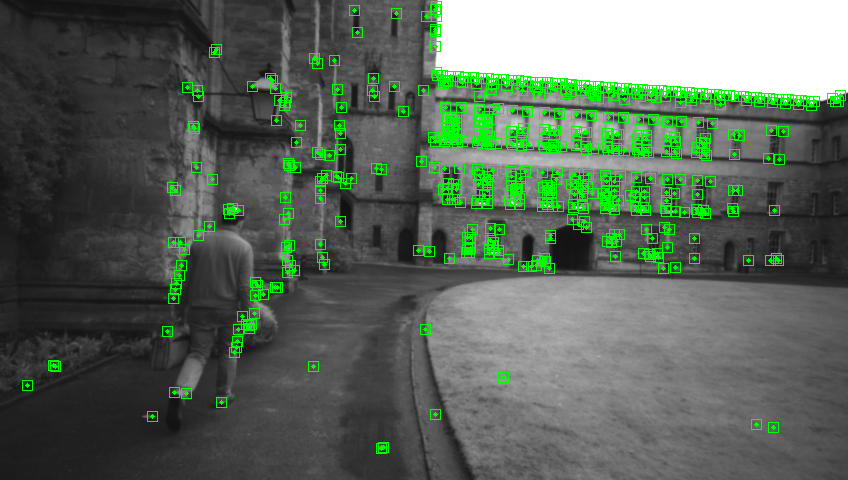}
        \end{minipage}
        
        \caption{}
        \label{Figure:orbslam3_fail_newer}
    \end{subfigure}
    
    \caption{(a) Variations in the RPE of ORB-SLAM3 when applied to the Newer College dataset and (b) associated frames with high RPE where most close points are extracted from a dynamic person.}
    \label{Figure:orbslam3_newer_combined}
\end{figure}

Since \cite{newercollege} has thoroughly analyzed the performance of different SLAM algorithms in large-scale environments in their source paper, we use the Newer College Quad and Mid section (the first 300 seconds of the short experiment sequence) in our experiment to discover more subtle differences with our tool. Stereo-based ORB-SLAM3 exhibits significant fluctuations in RPE. These fluctuations occur when a walking individual is visible in the frame, and the RPE stabilizes as the individual moves away from the camera and disappears, as demonstrated in Figures \ref{Figure:orbslam3_rpe_sub} and \ref{Figure:orbslam3_fail_newer}. ORB-SLAM3 relies on close points (those within a depth of 40 times the camera baseline) to offer more precise depth measurements compared to far points. However, as depicted in Figure \ref{Figure:orbslam3_fail_newer}, the majority of these close points are extracted from an unreliable dynamic person in frames with high RPE. Thus, ORB-SLAM3 is not accurate for dynamic environments. In contrast, LiDAR-based methods demonstrate commendable accuracy (see Table \ref{Table:lidar_comparison}).
Specifically, KISS-ICP is the most accurate system, while LeGO-LOAM, LOAM and F-LOAM show slightly worse accuracy. Note that by segmenting the ground points, LeGO-LOAM is able to filter out the noise from grass and leaves \cite{legoloam}, whereas LOAM and F-LOAM cannot.

\subsubsection{DARPA Subterranean Challenge}
\begin{figure}[!htb]
    \centering
    \begin{subfigure}[b]{0.49\columnwidth}
        \centering
        \includegraphics[width=\linewidth, height=0.7\linewidth]{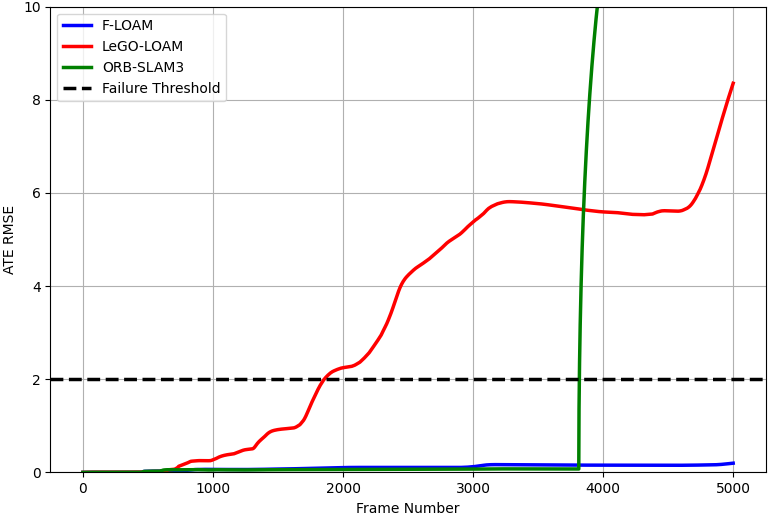}
        \caption{}
        \label{fig:darpa error compare}
    \end{subfigure}
    \hfill 
    \begin{subfigure}[b]{0.49\columnwidth}
        \centering
        \includegraphics[width=\linewidth, height=0.7\linewidth]{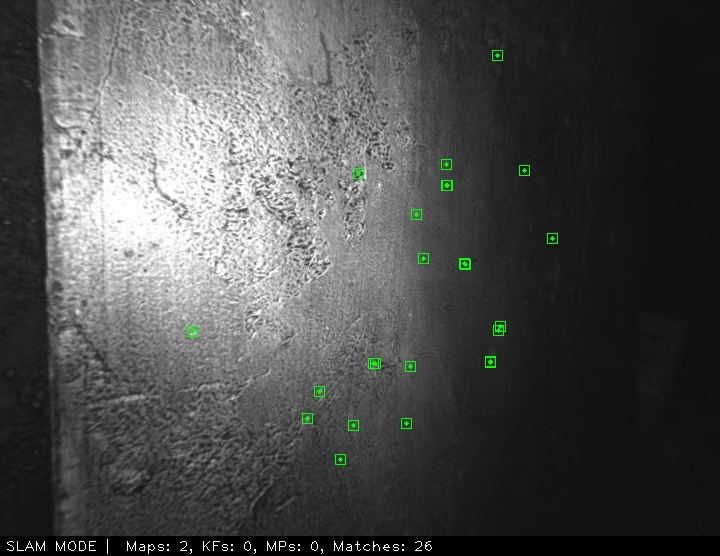}
        \caption{}
        \label{fig:orbslam3 darpa fail}
    \end{subfigure}
    
    \vspace{1ex} 
    
    \begin{subfigure}[b]{\columnwidth}
        \centering
        \includegraphics[width=0.97\linewidth]{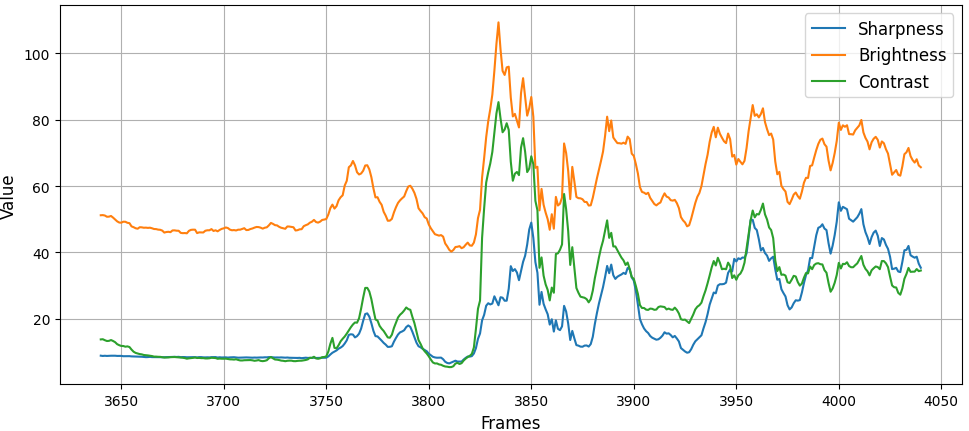}
        \caption{}
        \label{fig:darpa images quality}
    \end{subfigure}
    
    \caption{(a) Error comparison of ORB-SLAM3, LeGO-LOAM, and F-LOAM with a failure threshold of 2m; (b) Frame where ORB-SLAM3 fails to track local map in DARPA Subterranean Challenge; (c) Image metrics of previous and after 200 frames of ORB-SLAM3 failure point.}
    \label{fig:orbslam3_darpa_fail_combine}
\end{figure}

This dataset presents the most challenging scenarios, as three algorithms failed to build a consistent map; see Table \ref{Table:lidar_comparison}. ORB-SLAM3 consistently shows tracking failure of the local map at analogous places (see Figure~\ref{fig:orbslam3 darpa fail}). These failures are due to extreme lighting conditions, leading to inadequate extraction of ORB features. Note the increase in brightness and contrast in Figure~\ref{fig:darpa images quality}. LeGO-LOAM and KISS-ICP also encountered issues caused by sensor degradation within the subterranean tunnel, while LOAM and F-LOAM overcame this. It is worth noting that ORB-SLAM3 performs well before it encounters the failure point, while LeGO-LOAM starts to fail at the beginning of the sequence, as shown in Figure~\ref{fig:darpa error compare}.

\begin{figure}[tb!]
  \centering
  
    \begin{subfigure}[a]{0.49\linewidth}
            \includegraphics[width=\linewidth]{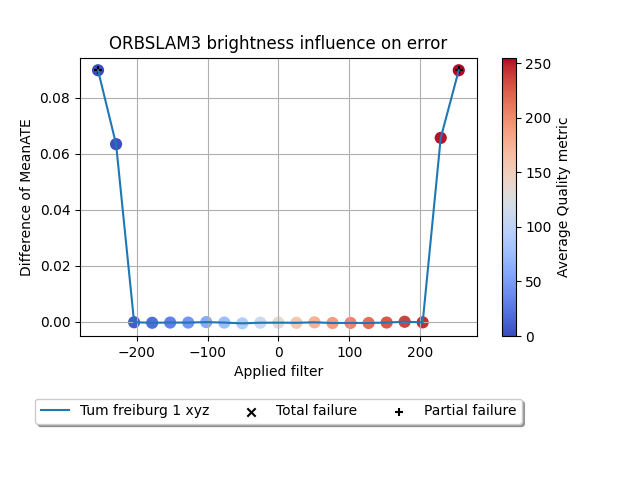}
    \end{subfigure}
    \begin{subfigure}[a]{0.49\linewidth}
            \includegraphics[width=\linewidth]{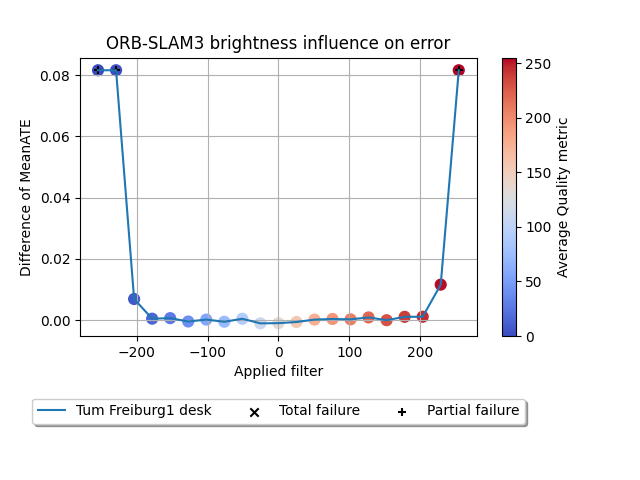}
    \end{subfigure}
    
    \begin{subfigure}[a]{0.49\linewidth}
        \includegraphics[width=\linewidth]{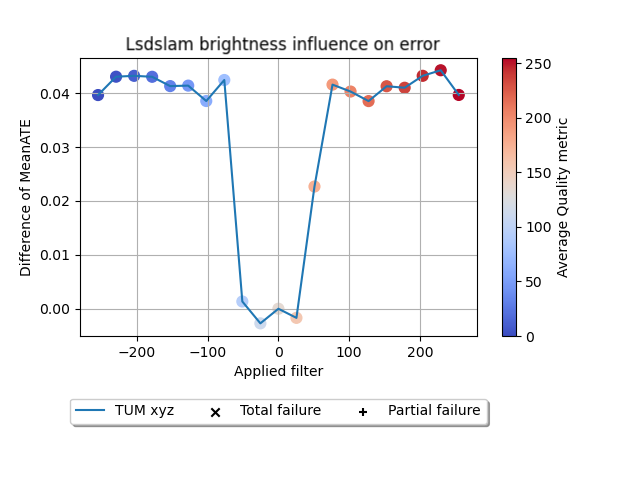}
    \end{subfigure}
    \begin{subfigure}[a]{0.49\linewidth}
        \includegraphics[width=\linewidth]{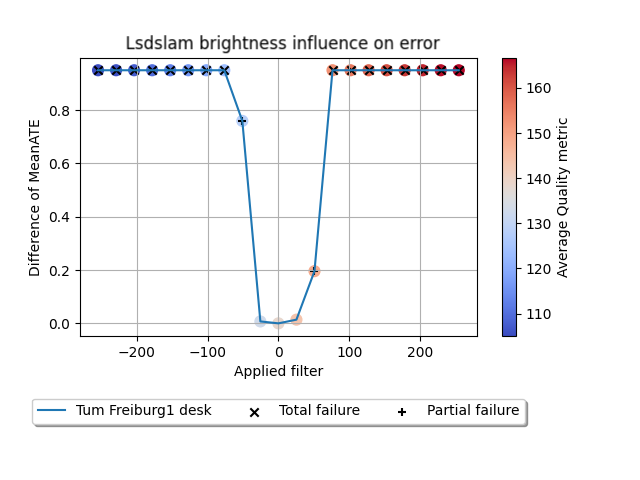}
    \end{subfigure}
    
    \begin{subfigure}[a]{0.49\linewidth}
        \includegraphics[width=\linewidth]{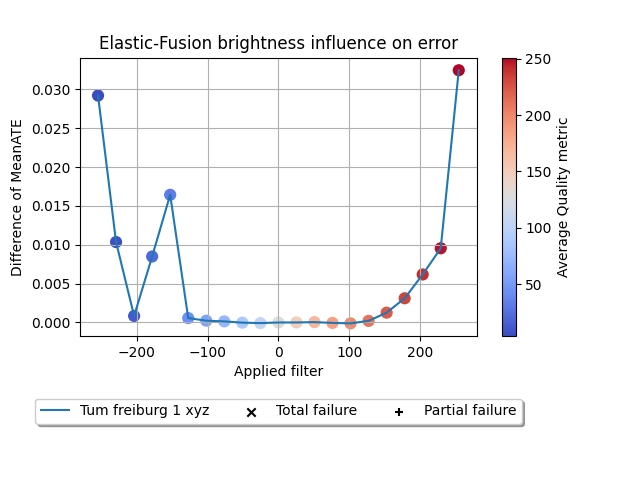}
    \end{subfigure}
    \begin{subfigure}[a]{0.49\linewidth}
        \includegraphics[width=\linewidth]{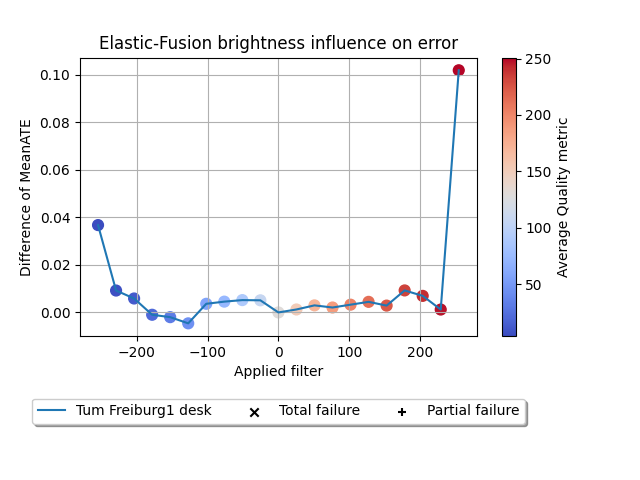}
    \end{subfigure}
    
    \caption{Brightness perturbation --- ORB-SLAM3 (top), LSD-SLAM (middle), ElasticFusion (bottom) on TUM Freiburg 1 xyz (left) and TUM Freiburg 1 Desk (right).}
    \label{fig:brightness}
\end{figure}

\subsection{SLAM Fuzzing}
%
\subsubsection{Image Quality and Breakpoints}
The quality of input data supplied to a visual SLAM algorithm impacts its performance. Variations in illumination, contrast levels, and image blur are known to affect algorithms differently. Understanding these influences can be challenging when benchmarking algorithms across different datasets, as it is often unclear whether observed behavioural differences stem solely from changes in brightness, contrast, or blur, or if other scene characteristics are at play. We use the fuzzing mechanism introduced in Section \ref{section:fuzzing}, to aid in this assessment. To illustrate its efficacy, we select three popular visual SLAM algorithms —  ORB-SLAM3, LSD-SLAM and ElasticFusion. We first select datasets where these algorithms perform well, and then we introduce incremental perturbations to random 1-second sequences, in total affecting up to 10\% of the trajectory. For contrast and brightness, the perturbation values range from -255 to 255 with increments of 25. For blur, the values range from 1 to 10, representing the size of the applied average blur kernel, with increments of 1. In the experiments, each perturbation is applied individually, but we can also explore how combinations of these perturbations affect algorithmic behaviour. For each applied perturbation value on each algorithm-trajectory pair, the experiments are repeated 5 times, with reported errors averaged. Partial failure refers to instances where the algorithm fails in one of the 5 runs due to reasons such as excessively high error, becoming stuck in one coordinate, or crashing outright. Total failure denotes scenarios where all 5 runs result in failure.

Figure \ref{fig:brightness} illustrates the impact of brightness perturbation on ORB-SLAM3, LSD-SLAM, and ElasticFusion. As anticipated, all graphs exhibit a general U shape, wherein both very dark and very bright images result in high errors. Of particular interest is the observation that the width of the curve can provide insights into the robustness of an algorithm, with LSD-SLAM exhibiting complete failure for lower perturbation values. Additionally, the results suggest that the effects of image brightness are somewhat dataset-dependent, as ORB-SLAM3 exhibits a slight initial error increase even with a relatively small change for TUM trajectories.

\begin{figure}[tb!]
    \centering

    \begin{subfigure}[a]{0.49\linewidth}
        \includegraphics[width=\linewidth]{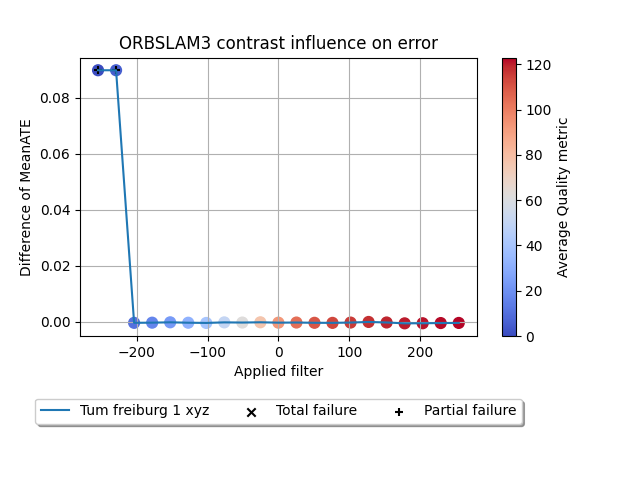}
    \end{subfigure}
    \begin{subfigure}[a]{0.49\linewidth}
        \includegraphics[width=\linewidth]{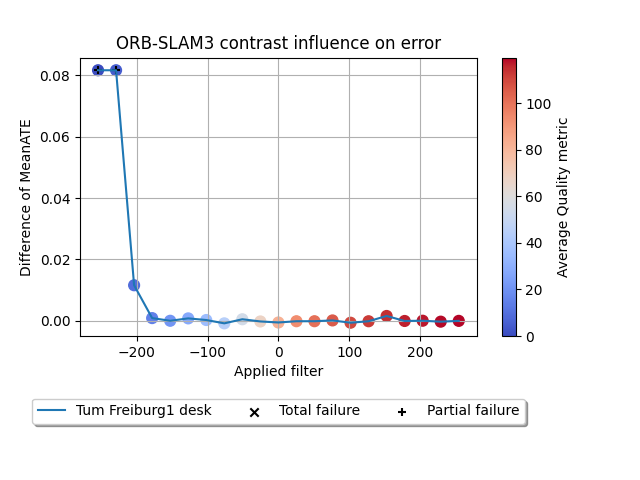}
    \end{subfigure}
    
    \begin{subfigure}[a]{0.49\linewidth}
        \includegraphics[width=\linewidth]{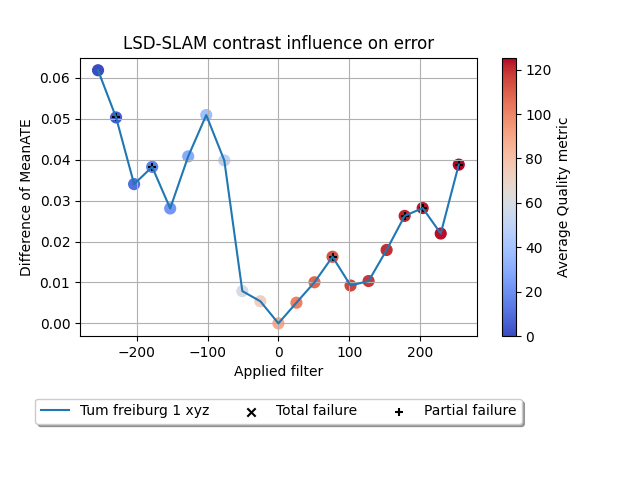}
    \end{subfigure}
    \begin{subfigure}[a]{0.49\linewidth}
        \includegraphics[width=\linewidth]{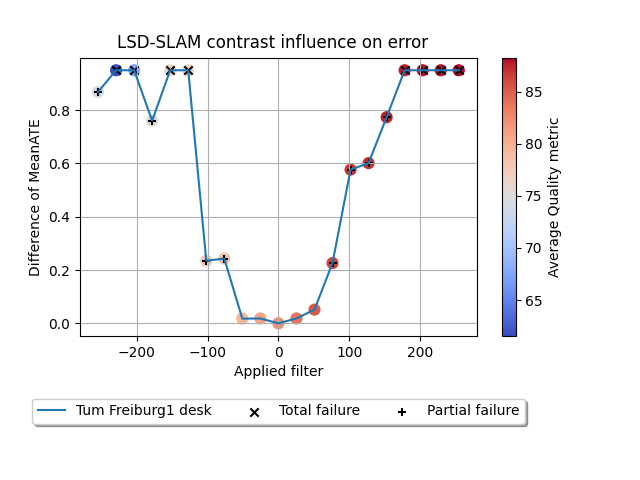}
    \end{subfigure}
    
    \begin{subfigure}[a]{0.49\linewidth}
        \includegraphics[width=\linewidth]{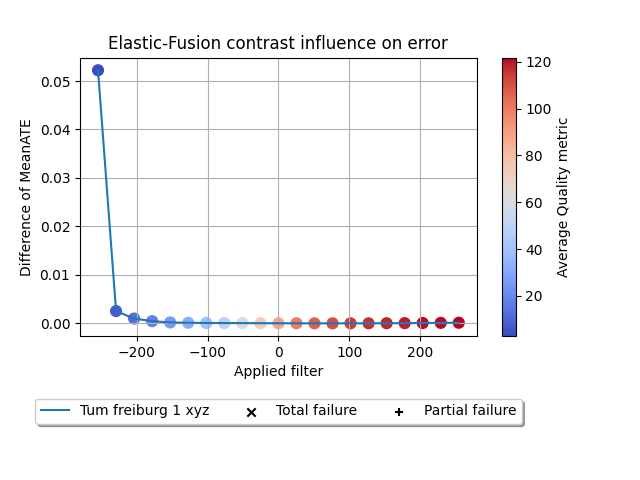}
    \end{subfigure}
    \begin{subfigure}[a]{0.49\linewidth}
        \includegraphics[width=\linewidth]{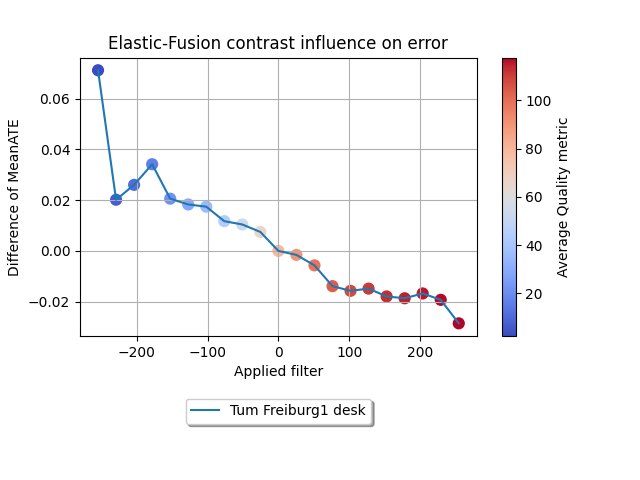}
    \end{subfigure}
    
    \caption{Contrast perturbation --- ORB-SLAM3 (top), LSD-SLAM (middle), ElasticFusion (bottom) on TUM Freiburg 1 xyz (left) and TUM Freiburg 1 Desk (right).}
    \label{Figure:orbslam3_contrast}
\end{figure}

\begin{figure}[!tb]
    \centering
    \begin{subfigure}[a]{\columnwidth}
       \begin{minipage}[a]{0.40\linewidth}
           \includegraphics[width=\linewidth]{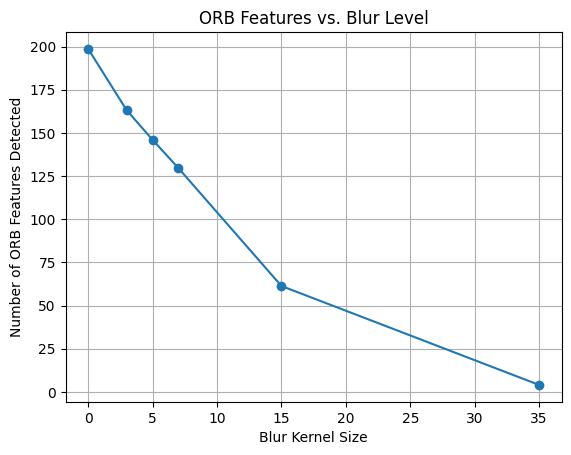}
       \end{minipage}
        \hfill
        \begin{minipage}[a]{0.55\linewidth}
            \includegraphics[width=\linewidth]{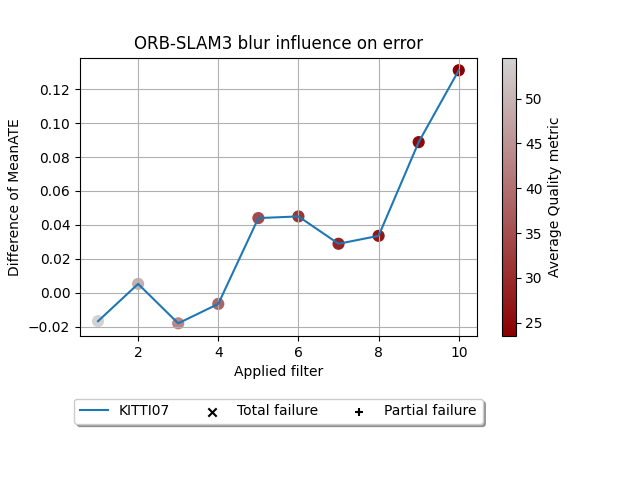}
        \end{minipage}
        
    \end{subfigure}

    \caption{Blur perturbation --- number of ORB features detected (left) and ORB-SLAM3 error (right) using KITTI.}
    \label{fig: blur}
\end{figure}

We can analyse whether certain metrics prioritize higher or lower values by assessing their impact on system performance or behaviour. For instance, we observe varying responses across algorithms when perturbing contrast values. Unlike brightness, where a consistent trend was observed, contrast alterations resulted in diverse outcomes. ORB-SLAM3, for instance, appears to be more sensitive to low-contrast data, exhibiting greater resilience to high-contrast variations (Figure \ref{Figure:orbslam3_contrast}). 
High-contrast data seems to preserve the edges of the original image. 
Conversely, LSD-SLAM, while more resilient to contrast changes than brightness variations, appears to be affected by alterations in both low and high contrast (Figure \ref{Figure:orbslam3_contrast}). Some of the experiments reveal unexpected situations. For example, Figure \ref{Figure:orbslam3_contrast} shows that increasing contrast does not degrade  ElasticFusion, but rather makes it temporarily improve ATE by 1\% compared to the baseline.

Concerning blur, the experiments reveal a consistent trend across all tested algorithms, characterized by a gradual decline in performance as blur increases.When features are blurred, they become harder to detect and match between frames, leading to tracking failures and increased positional errors.  For instance, as depicted in Figure \ref{fig: blur}, ORB-SLAM3 exhibits an ascending error with blur increase, likely due to the reduced number of ORB features identified with increasing blur. As the algorithm depends heavily on these ORB features to maintain its accuracy, the diminishing quality of input data directly impacts its performance, resulting in an increase in error.

\subsubsection{Image Quality and Loop Closure}
Unlike LiDAR-based SLAM, loop closure in visual SLAM is influenced by various factors beyond geometry, such as lighting conditions. The introduced fuzzing tool tests the loop closure robustness of ORB-SLAM3 which relies on the Bag of Words method.
We first identify \( n \) pairs of frames \( (a; b)_n \) that can form loop connections without extra perturbation. For each frame, we then group the preceding and subsequent 30 continuous frames to obtain \( (A = a_{i-30}, \ldots, a_{i+30}; B = b_{j-30}, \ldots, b_{j+30})_n \). For each pair, we introduce perturbations to \( A \) or \( B \) until no frame in \( A \) can form a loop connection with any frame in \( B \). We record the maximum perturbation for which a loop connection is maintained for each pair, calculate the corresponding percentage difference in image metrics of the connected frames, and average across all evaluated loop closures to obtain an estimated threshold value for different image metrics, as shown in Table~\ref{Table:loop_closure_image}. It suggests that two images taken from the same location are likely to form a loop connection if the percentage difference in various image metrics is below this threshold.

\begin{table}[tb!]
\caption{Estimated Threshold of Percentage Difference in Image Metrics for Correct ORB-SLAM3 Loop Closure}
\centering
\begin{tabularx}{\columnwidth}{|C|C|C|} 
\hline
Tenengrad & Brightness & Contrast \\
\hline
50.4\%    & 59.7\%     & 54.1\%   \\
\hline
\end{tabularx}
\label{Table:loop_closure_image}
\end{table}

\begin{figure}[tb!]
    \centering
    \includegraphics[width=0.9\columnwidth]{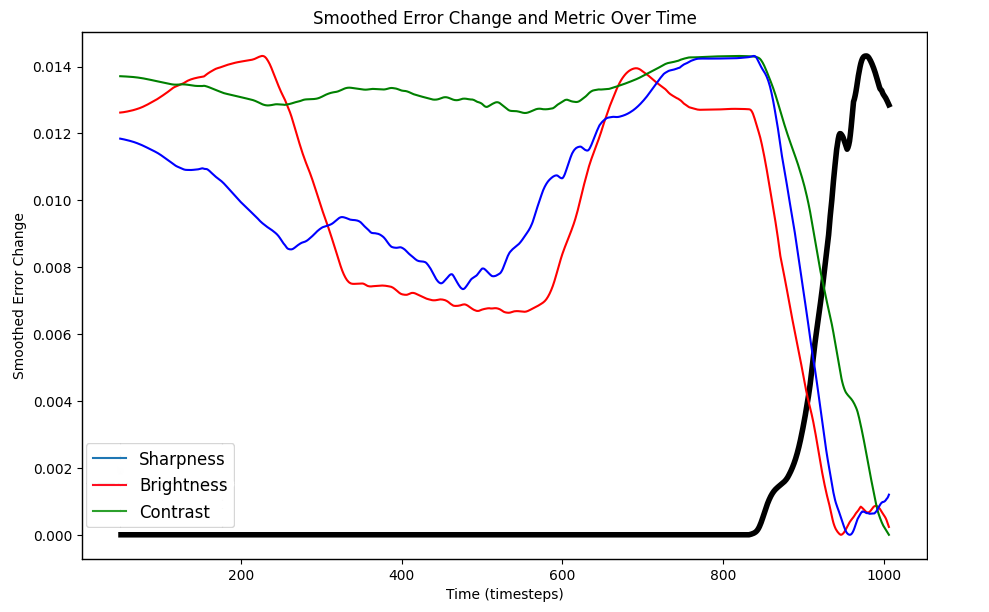}
    \caption{Change in error (black) and image metrics (red, green and blue) over time with error increasing sharply after frame 800 when drone takes off for LSD-SLAM on the \textit{indoor\_forward\_7\_snapdragon} sequence of UZH-FPV.}
    \label{Figure:lsd-drone}
\end{figure}

\subsection{Performance Diagnostics}

As described in Section \ref{peformance}, our tool facilitates the identification of frames of interest to investigate the relationships between input data and algorithm behaviour. To illustrate this, we conduct experiments with  LSD-SLAM on the UZH-FPV dataset, specifically utilizing the indoor forward-facing trajectory. To identify frames of interest -- where significant changes in error occur -- we generate plots illustrating the correlation between our visual image metrics and the error increase. Figure \ref{Figure:lsd-drone} displays one such plot for a part of LSD-SLAM, corresponding to the moment when the drone takes off. During takeoff, sharpness, brightness, and contrast drop, while the error increases drastically. We conclude that the analysed image metrics can serve as good diagnostic indicators, albeit not definitive predictors of failure. 
While they provide valuable insights into the quality of visual data and help identify where the failure occurred, specific values do not guarantee failure. Instead, our approach focuses on utilizing them to gain insights into the underlying causes of performance degradation. Nonetheless, one limitation of relying solely on image metrics is the potential oversight of other causes of failure, such as occlusions or sensor malfunctions. Therefore, a good understanding of SLAM performance and failure modes may require considering causes beyond those evaluated by the implemented metrics.

\section{Conclusions}
This paper has introduced SLAMFuse, an open-source framework designed for the rigorous benchmarking and performance diagnosis of SLAM systems. Central to SLAMFuse is its innovative use of Docker containers and volumes which aims to resolve prevalent software maintenance issues associated with library dependencies.

Furthermore, SLAMFuse has the distinguishing feature of evaluating SLAM algorithms across a variety of sensory modalities. This is achieved through the integration of state-of-the-art LiDAR algorithms and the inclusion of challenging datasets designed to test the algorithms under a wide range of conditions. A key contribution of SLAMFuse is its novel "fuzzing" mechanism, which systematically introduces perturbations into dataset inputs to stress test SLAM algorithms against unexpected environmental changes. SLAM-fuzzing, along with advanced diagnostic tools, facilitates detailed, frame-by-frame analysis of algorithmic performance and error conditions.

The Docker technology introduces a small overhead, quantitatively assessed to range between 5\% and 10\% in terms of execution performance. However, this overhead is justified by the significant advantages offered by SLAMFuse, particularly in terms of enhancing the consistency and portability of SLAM algorithm evaluation.

Through extensive experimentation, utilizing a diverse array of datasets and scenarios, SLAMFuse has proven to be an invaluable asset in the performance analysis of SLAM algorithms. The SLAMfuse framework empowers researchers under the ``microscope'' of current SLAM technologies.

\section*{Acknowledgements}
This work is partially funded by EPSRC EP/N035127/1 (LAMBDA project) and EP/T026995/1 (EnnCore project) and by the UK Industrial Strategy Challenge Fund (ISCF) under the Digital Security by Design (DSbD) Programme delivered by UKRI as part of the Soteria (75243) project. Mikel Luj\'an is supported by a Royal Society Wolfson Fellowship and an Arm/RAEng Research Chair Award.


\bibliographystyle{IEEEtran}
\bibliography{bibliography}

\end{document}